\documentclass[journal,twoside,web]{ieeecolor}
\usepackage{generic}
\usepackage{cite}
\usepackage{amsmath,amssymb,amsfonts}
\usepackage{graphicx}
\usepackage{epstopdf-base}
\usepackage{hyperref}
\hypersetup{hidelinks=true}
\usepackage{textcomp}
\usepackage{algorithm}
\usepackage{algpseudocode}
\usepackage{algorithmicx}%
\usepackage{algpseudocode}%
\usepackage{listings}%
\usepackage{tabularx} %
\makeatletter
\let\labelindent\relax
\makeatother
\usepackage{enumitem}

\usepackage{xcolor}%
\usepackage{siunitx}
\usepackage{multirow}
\usepackage{nicematrix}
\usepackage{float}
\usepackage{rotating}

\usepackage{placeins}
\usepackage{stfloats}   

\providecommand{\refname}{References}

\usepackage[capitalise,nameinlink]{cleveref}
\crefname{figure}{Fig.}{Figs.}
\Crefname{figure}{Fig.}{Figs.}

\crefname{table}{Table}{Tables}
\Crefname{table}{Table}{Tables}

\crefname{section}{Section}{Sections}
\Crefname{section}{Section}{Sections}

\crefname{subsection}{Section}{Sections}
\Crefname{subsection}{Section}{Sections}

\crefname{equation}{Eq.}{Eqs.}
\Crefname{equation}{Eq.}{Eqs.}

\crefname{algorithm}{Algorithm}{Algorithms}
\Crefname{algorithm}{Algorithm}{Algorithms}

\definecolor{octnweb}{rgb}{0,0.263,0.576}
\newcommand{\projectwebsite}[1]{%
  \\[0.45em]\centerline{\normalsize\sf\color{octnweb}%
  \href{https://#1}{\underline{Project Website}}}}

\long\def\rp#1{{\color{black}#1}}
\long\def\pr#1{{\color{black}#1}}

\def\BibTeX{{\rm B\kern-.05em{\sc i\kern-.025em b}\kern-.08em
    T\kern-.1667em\lower.7ex\hbox{E}\kern-.125emX}}
\begin{document}
\title{OCTN: Neural OCT Representations for Robot-Guided Precision Intervention}
\author{Ravi Prakash, \IEEEmembership{Student Member, IEEE}, Ryan P. McNabb \IEEEmembership{Member, IEEE}, Patrick J. Codd, and Shan Lin
\projectwebsite{raprakashvi.github.io/octn}
\thanks{This work has been submitted to the IEEE for possible publication. Copyright may be transferred without notice, after which this version may no longer be accessible. \emph{Manuscript currently under review.}}
\thanks{Aspects of the technology described in this manuscript are the subject of the following patent application: R. Prakash, R. P. McNabb, P. J. Codd, and S. Lin, ``Systems and methods for neural OCT tissue representations for robot-guided intervention,'' U.S.\ Provisional Patent Application No.~64/045,412, patent pending.}
\thanks{Corresponding author: Ravi Prakash.}
\thanks{Ravi Prakash, Ryan P. McNabb, and Patrick J. Codd are with Duke University, Durham, NC 27705 USA (e-mail: firstname.lastname@duke.edu).}
\thanks{Shan Lin is with Arizona State University, Phoenix, AZ 85287 USA (e-mail: shan.lin.2@asu.edu).}
}

\maketitle

\begin{abstract}
Optical coherence tomography (OCT) offers compact, contactless, micron-scale imaging suitable for intraoperative guidance, but native OCT volumes are discretely sampled, anisotropic, and \pr{currently} inefficient for downstream geometric reasoning and robot integration. %
We present OCTN (pronounced ``octane”), an implicit neural representation framework that converts volumetric OCT scans into a continuous, differentiable, and spatially faithful tissue-intensity field. OCTN uses a two-stage hybrid training strategy that combines supervision from acquired voxels with inter-slice interpolations, preserving B-scan fidelity while improving continuity in sparsely sampled regions. For versatility, we first show that OCTN enables fast volumetric reasoning by storing the learned tissue representation natively on the GPU, \pr{supporting} intensity-based spatial queries with up to $43\times$ speedup over conventional CPU processing. %
We then demonstrate OCTN-\pr{enabled} OCT-guided robotic laser surgery \pr{where the continuous tissue representation supports implicit surface discovery and surface-constrained path planning via multiple optimization strategies, including Newton- and SGD-based optimization.} 
Next, OCTN enables reconstruction of dense volumetric structure from sparsely acquired B-scans, while reducing acquisition time by $4\times$ and preserving \pr{clinically} relevant structures. Across the newly generated \textit{Duke TissueOCT} dataset and public OCT datasets, OCTN achieves \pr{robust} high-fidelity reconstruction with PSNR $>30$~dB with training time $<10$~s and \pr{preserves surface consistency} within $10\mu$m Chamfer distance relative to baseline reconstruction. The TissueOCT dataset and code are publicly available \pr{at raprakashvi.github.io/octn}. 
\end{abstract}


\section{Introduction}
\label{sec:introduction}

 \begin{figure*}[!tb]
    \centering
    \includegraphics[width=0.9\linewidth]{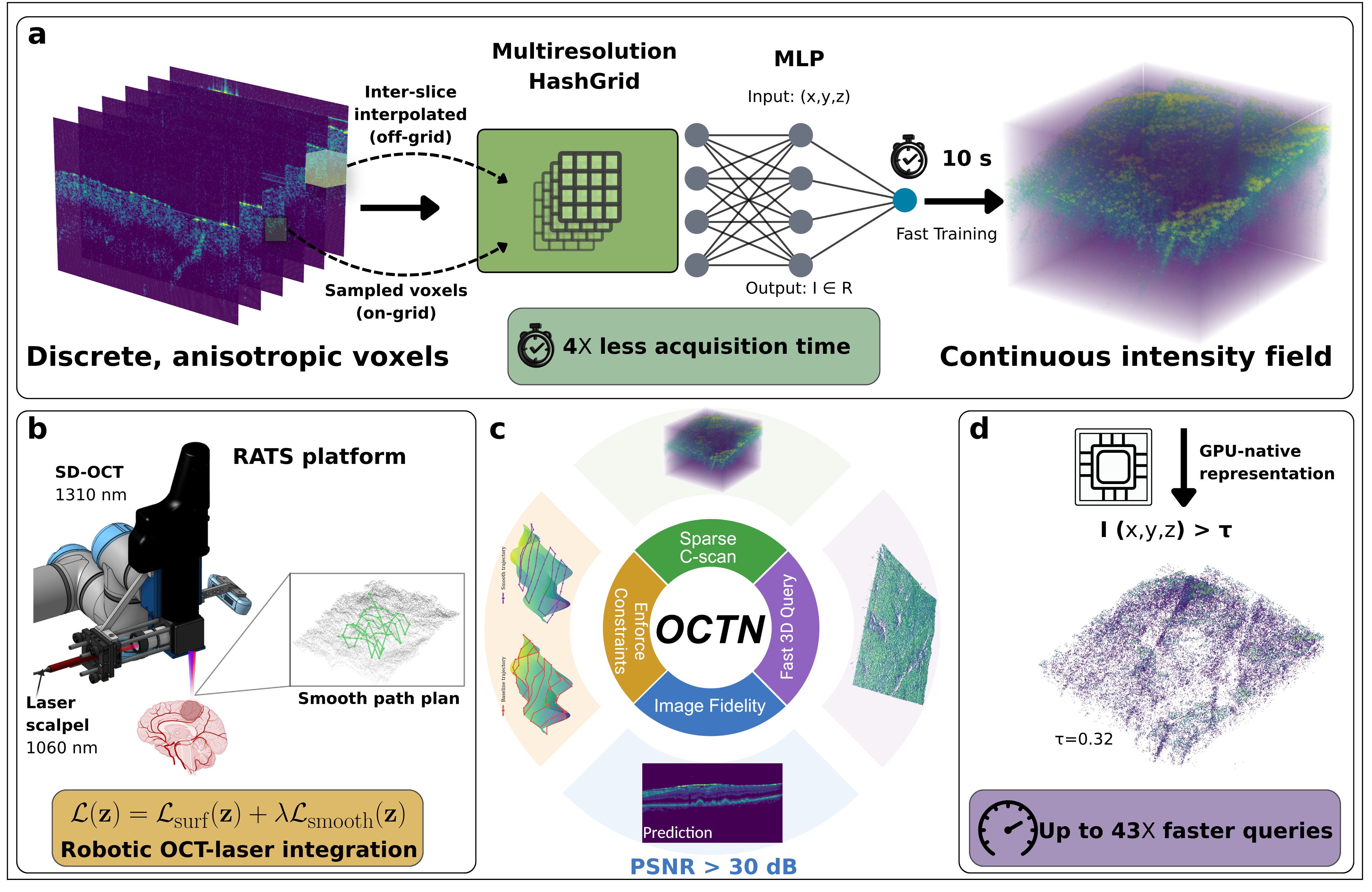}
\caption{\textbf{System overview.} 
\textbf{a)} OCTN maps discrete, anisotropic OCT voxels to a continuous tissue-intensity field using on-grid samples and off-grid inter-slice supervision. Its multiresolution hash-grid and MLP enable fast training, arbitrary spatial queries, and $4\times$ sparse acquisition.
\textbf{b)} OCTN supports smooth, surface-aware robotic laser path planning. 
\textbf{c)} OCTN balances sparse C-scan reconstruction, image fidelity, fast 3D querying, and surface constraints. 
\textbf{d)} Its GPU-native representation enables threshold-based tissue queries with up to $43\times$ faster spatial reasoning.}
    \label{fig:system_overview}
\end{figure*}

\IEEEPARstart{O}{ptical} Coherence Tomography (OCT) is a contactless, non-ionizing sensing modality suited for high-resolution, near-real-time surgical imaging \cite{huang1991optical}. OCT allows micron-scale visualization of surface and shallow subsurface tissue features within a compact form factor. These properties make OCT attractive for intraoperative imaging and robot-guided intervention \cite{draelos2021contactless} over conventional counterparts. Although currently intraoperative ultrasound (IOUS) is widely used, it requires acoustic contact and provides limited spatial resolution ($\sim0.5\,\mathrm{mm}$) compared to OCT. %
Since its inception, OCT has established clinical use in ophthalmology \cite{toth1997comparison,wojtkowski2004ophthalmic,tan2018overview} and has more recently been explored in neurosurgery \cite{kut2015detection, dolganova2020capability,navabi2025computer}, breast imaging \cite{boppart2004optical, bareja2022classifying}, dermatology \cite{dey2022skin, latriglia2023line}, and urology \cite{ma2024large}. However, broader adoption remains limited by a restricted field of view (FoV), anisotropic scaling, and a computationally cumbersome discrete volume representation.

Recent advances in robot-mounted OCT \cite{ma2024large,prakash2025portable, ma2025geometry,ma2025robotics,lotz2024large,prakash2025see} have enabled a larger field of view (FoV), surface-aware high-resolution imaging \cite{fang2025robotic}, extending utility. Despite this progress, native OCT volumes remain discretely sampled with anisotropic resolution and sampling, making them cumbersome to use directly for downstream computation. As a result, geometric reasoning and planning over OCT data remain inefficient and slow. 
Therefore, devising an isotropic and queryable representation of OCT data can overcome these limitations. 
Implicit neural representation (INR) is a Multi-Layer Perceptron (MLP) based method to parameterize the tissue volume by learning a mapping from spatial coordinates to intensity \cite {molaei2023implicit}. This results in a continuous, isotropic,  and queryable neural encoding of the volume.

In recent years, INR have gained prominence in medical imaging \cite{molaei2023implicit} for reconstruction \cite{li2025computational, saha2025based}, super-resolution \cite{chen2023cunerf, kahrs2026don}, segmentation, and registration \cite{kahrs2026don, komeda2026high}. Unlike INR for natural 3D scenes, which often rely on multi-view image capture \cite{liu2024baa}, medical imaging modalities are typically reconstructed from sparsely sampled parallel single-view cross-sectional slices. Methods such as CuNeRF \cite{chen2023cunerf} have adapted INRs for CT and MRI scans by learning a mapping from 3D spatial coordinates to voxel intensity, yielding a spatially correct continuous volume representation. When available, auxiliary information such as attenuation can further augment input representation and enhance reconstruction quality \cite{zhou2024rho}. Expanding these methods to OCT requires handling sensor-specific idiosyncrasies and anisotropies under tight computational constraints for relatively small tissue volumes. 
Prior works have addressed OCT sampling anisotropy either through additional sensing modalities such as scanning laser ophthalmoscopy \cite{kahrs2026don} or through physics-informed modeling of A-scan formation \cite{li2025computational}. Despite these advances, existing INR-based OCT methods rely on specialized hardware, multimodal sensing, and prolonged training time, limiting their practicality for broader intraoperative deployment.

In this work, we present \textbf{OCTN}, an INR-based method for learning a continuous representation of OCT volumes under anisotropic sampling and speckle noise (\cref{fig:system_overview}a). \pr{OCTN enables constraint-based optimization for robot path planning, \cref{fig:system_overview}b, with high-fidelity image reconstruction \cref{fig:system_overview}c, supports fast volumetric queries \cref{fig:system_overview}d, while recovering spatially accurate geometry from a sparsely sampled volume.} \pr{OCTN enhances the existing OCT experience in terms of resolution, speed, and robotic guidance.} 
%
Overall, we propose:
\begin{itemize}[leftmargin=*, itemsep=0.2em]

    \item \textbf{Hybrid framework for continuous OCT representation:} We introduce a two-stage training strategy combining discrete and interpolated voxels to learn a continuous, isotropic OCT representation. 
     \item \textbf{Fast query and robot path planning:} We demonstrate OCTN enables $19\times$--$43\times$ faster volumetric queries than conventional CPU-based processing through a GPU-native implementation. Furthermore, its continuous representation supports constrained surface-aware path planning using Newton- and Stochastic Gradient Descent (SGD)-based optimization \pr{separately}, on an OCT-guided robotic laser platform \cite{prakash2025see} \cref{fig:system_overview}b.
    \item \textbf{Sparse volume reconstruction:} We show that OCTN supports reconstruction from sparsely sampled B-scan stacks, enabling up to $4\times$ reduction in B-scan density within clinical constraints. For a critically sampled volume, this corresponded to reducing acquisition time from $24$~s to $6$~s.
     \item \textbf{\pr{TissueOCT Benchmarking}:} We release an open-source \pr{Duke TissueOCT} dataset \pr{and comprehensive set of global and local metrics to benchmark learned B-scan fidelity, sparse to dense volume reconstruction quality, visualization speed for volumetric queries, and surface aware path planning over realistic tissue samples.} \pr{The dataset consists of} critically sampled OCT volumes from \textit{ex vivo} tissues at varying noise and artifact levels, \pr{as seen in \cref{tab:dataset_summary}}, together with calibration parameters \pr{for 3D reconstruction.}  
\end{itemize}

\begin{figure*}[!tb]
    \centering
    \includegraphics[width=\textwidth]{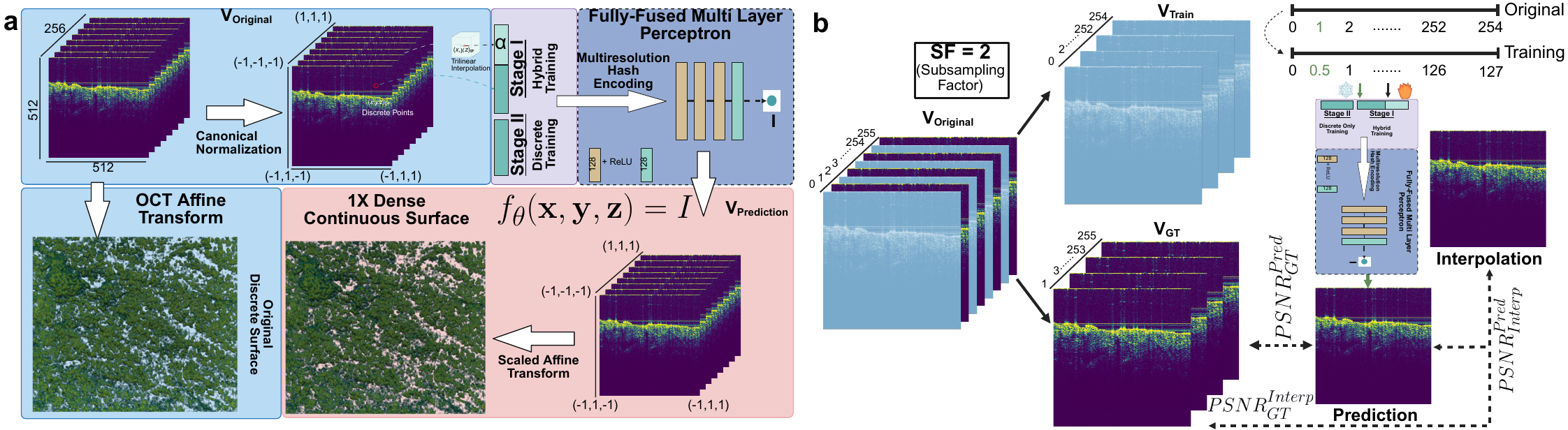}

    \caption{\textbf{OCTN Framework and sparse reconstruction}. \textbf{a}) OCTN converts an anisotropically sampled OCT C-scan, $V_{\mathrm{Original}}$, into a continuous neural volumetric representation. Volume is mapped to a canonical coordinate domain $[-1,1]^3$, and an INR $f_\theta(x,y,z)\rightarrow I$ is trained using a two-stage strategy. Stage I combines acquired discrete voxels with interpolated inter-slice in ratio ($\alpha$) to encourage off-grid continuity, while Stage II performs discrete-only refinement. The learned field can be queried at arbitrary resolution, enabling dense surface extraction and continuous volume rendering. %
    \textbf{b}) Sparse reconstruction protocol shown for $SF=2$. The subsampled training volume, $V_{\mathrm{Train}}$, is used to train OCTN, and held-out B-scans form $V_{\mathrm{GT}}$ for evaluation. OCTN predictions, $V_{\mathrm{Pred}}$, are compared with ground truth and with trilinear interpolation (baseline) to quantify reconstruction quality at unobserved slice locations. }
    \label{fig:model_architecture}
\end{figure*}

\section{Materials and Methods}
\label{sec:methods}
Here we describe our core methods and evaluation pipelines. This section is broadly divided into three broad themes: \textbf{i)} OCTN Core (dataset \cref{sec:Data_acquisition_datasets}, OCTN framework \cref{sec:OCTN framework}, training pipeline \cref{sec:Model_Training}, \& speed optimization \cref{sec:gpu_optimization_for_inference}), \textbf{ii)} evaluation schemes and protocols (evaluation metrics \cref{sec:eval_metrics} and sparse volume reconstruction \cref{sec:intermediate_and_sparse_slice_eval_protocol}), \textbf{iii)} downstream robotic laser based tissue ablation by integrating OCTN for path planning \cref{sec:surface_constrained_depth_projection_method}).

\subsection{Duke TissueOCT dataset} \label{sec:Data_acquisition_datasets}
We present the \textbf{Duke TissueOCT dataset} to benchmark OCTN performance for volumetric reconstruction, sparse acquisition, and 3D tissue representation. %
The dataset, as described in \cref{tab:dataset_summary}, contains OCT volumes of heterogeneous \textit{ex vivo} animal tissues.
Data were acquired using a low-cost spectral-domain (SD)-OCT system with a $1310$~nm central wavelength and a $110$~mm focal length (Lumedica Inc, NC, United States). Each rectangular volume had dimensions of $[D,H,W]=[256,512,512]$, where $D$ is the number of B-scans and $H,W$ denote the B-scan height and width. The voxel spacing was $14.59\,\mu\mathrm{m}$ axially, $14\,\mu\mathrm{m}$ laterally, and $28\,\mu\mathrm{m}$ between adjacent B-scans, corresponding to a critically sampled volume spanning $7.17 \times 7.17 \times 7.47~\mathrm{mm}^3$ with tissue-dependent axial penetration of $2$--$4$~mm. %
Porcine muscle samples were acquired under two noise-suppression settings, with a subset containing artificial dye artifacts from cancer tissue-marking dye. Bovine and chicken tissue samples were acquired under a low-speckle-noise setting. \pr{As INR is susceptible to noise and prone to overfitting to easy-to-learn structures, the TissueOCT dataset contains a diversity in noise level, tissue type, and optical contrast to pose it as fair benchmarking. }

\subsection{{OCTN: Continuous Neural Representation of Volumes}} \label{sec:OCTN framework}

Given an OCT C-scan, $\mathbf{[D, H, W]}$, OCTN learns a continuous neural intensity field that maps spatial coordinates to voxel intensity. Voxel intensities are normalized between $[0,1]$, and coordinates are mapped to the canonical frame $[-1,1]^3$, providing a resolution agnostic representation across volumes. OCTN uses a two-stage training strategy comprised of a hybrid and discrete training phase, further detailed in \cref{sec:Model_Training}. The hybrid phase learns an isotropic continuous representation for smooth sub-voxel queries by utilizing a mixture of discrete voxels from the acquired volume and interpolated counterparts generated through trilinear interpolation during training time, \cref{fig:model_architecture}a. During the later discrete training phase, the emphasis is on assuring high-fidelity reconstruction on the native OCT grid. \pr{Having a fraction of interpolated voxels mixed with discrete counterparts prevents overfitting on native voxels, and improves overall robustness of representation.} The resultant neural tissue representation is given by \cref{eq:neural_field_basic}.
%
\begin{equation}
f_{\theta}(x,y,z) = I
\label{eq:neural_field_basic}
\end{equation}
where $(x,y,z)$ denotes a canonical 3D coordinate and $I$ denotes the predicted OCT intensity. Unlike a discrete voxel array, the formulation supports intensity evaluation at arbitrary spatial coordinates, including unobserved locations between acquired B-scans.

\begin{table}[!tb]
\centering
\caption{Duke TissueOCT sample distribution.}
\label{tab:dataset_summary}
\scriptsize
\setlength{\tabcolsep}{3pt}
\renewcommand{\arraystretch}{1.02}
\begin{tabularx}{\columnwidth}{Xc}
\hline
\textbf{Tissue} & \textbf{Volumes} \\
\hline
Porcine muscle, high SNR & 5 \\
Porcine muscle, low SNR & 11 \\
Chicken breast & 12 \\
Bovine tissue & 11 \\
Porcine muscle with dye & 7 \\
\hline
\textbf{Total} & \textbf{46} \\
\hline
\end{tabularx}
\end{table}

\begin{algorithm}[!t]
\caption{OCTN training procedure}
\label{alg:training}
\begin{algorithmic}[1]
\Require OCT volume $\mathbf{V}$, neural field $f_{\theta}$, iterations $T$, hybrid iterations $T_{\mathrm{hyb}}$, batch size $B$, fraction $\alpha$
\State Initialize $\theta$, optimizer, and learning-rate scheduler
\For{$t = 1$ to $T$}
    \If{$t \leq T_{\mathrm{hyb}}$}
        \State $C_{\mathrm{interp}} \gets \lfloor \alpha B \rfloor$, $C_{\mathrm{disc}} \gets B - C_{\mathrm{interp}}$
        \State Sample $\mathbf{c}_{\mathrm{interp}} \sim \mathcal{U}([-1,1]^3)$ and $\mathbf{c}_{\mathrm{disc}}$ from the native grid
        \State Get $I_{\mathrm{interp}}$ by trilinear interpolation and $I_{\mathrm{disc}}$ by voxel lookup in $\mathbf{V}$
        \State $\mathbf{c} \gets \mathbf{c}_{\mathrm{interp}} \cup \mathbf{c}_{\mathrm{disc}}$, $I \gets I_{\mathrm{interp}} \cup I_{\mathrm{disc}}$
    \Else
        \State Sample $\mathbf{c}_{\mathrm{disc}}$ from the native grid and obtain $I_{\mathrm{disc}}$ by voxel lookup
        \State $\mathbf{c} \gets \mathbf{c}_{\mathrm{disc}}$, $I \gets I_{\mathrm{disc}}$
    \EndIf
    \State $\hat{I} \gets f_{\theta}(\mathbf{c})$
    \State $\mathcal{L}_{\mathrm{rec}} \gets \mathcal{L}(\hat{I}, I)$
    \State Update $\theta$ by backpropagation
\EndFor
\State \Return $f_{\theta}$
\end{algorithmic}
\end{algorithm}


\subsection{Hybrid Discrete-Interpolated Training } \label{sec:Model_Training}

First, a discrete OCT volume $\mathbf{V} \in \mathbb{R}^{D \times H \times W}$, is mapped to canonical coordinates through \cref{eq:coord_field}. 

\begin{equation}
x_i=\frac{2i}{W-1}-1,\;
y_j=\frac{2j}{H-1}-1,\;
z_k=\frac{2k}{D-1}-1 
\label{eq:coord_field}
\end{equation}

This is followed by the division of the training process into hybrid and discrete phases based on $r_{\mathrm{hyb}}$ and $r_{\mathrm{disc}}$ ratios for total iterations, respectively. For a batch of size $B$, $\alpha$ denotes the fraction of interpolated samples during the hybrid phase, detailed in \cref{alg:training}.
Each batch during the hybrid phase contains a sample distribution given by $C_{\mathrm{interp}} = \lfloor \alpha B \rfloor$ and $C_{\mathrm{disc}} = \lfloor B - C_{\mathrm{interp}} \rfloor$ respectively. \pr{With speed in mind, a fully fused multi-layer perceptron (MLP) is used to learn the volume \cite{muller2022instant, tiny-cuda-nn}. Voxel coordinates in canonical space undergo projection to high-dimensional space through multiresolution hash encoding to accurately represent high-frequency tissue features \cite{mueller2022instant}. Specifically, the hashgrid encoder used has 24 levels, with 2 $features/level$ and a hash table of size $2^{20}$ with base resolution of 16, following standard guidelines \cite{muller2022instant}. Encoded features are passed to MLP with 3 hidden layers with 128 neurons per layer and ReLU activation, mapped to a linear output head to regress intensity. The ADAM optimizer is used during training with mean squared error loss.}

\subsection{GPU-first Implementation} \label{sec:gpu_optimization_for_inference}

With Tiny-CUDA-NN as backbone, OCTN achieves orders of magnitude training speed gain over contemporary methods \cite{chen2023cunerf}, due to GPU-optimized kernels that fuse multiple operations into a single CUDA kernel. This removes memory bandwidth bottlenecks, allowing OCTN to run efficiently even on a standard NVIDIA GPU, expanding access, unlike popular methods \cite{chen2023cunerf,kahrs2026don}. OCTN has been successfully tested on NVIDIA 3070 Ti, 4080 Super, and 5090 GPUs.  
The result is a continuous neural intensity field,\cref{eq:neural_field_basic}, that maps canonical coordinates $(x,y,z) \in [-1,1]^3$ to voxel intensity $I$.  

All metrics and comparisons reported are for a single NVIDIA 4080 Super GPU. During training, the full OCT volume is loaded onto the GPU to minimize data transfer between the CPU and GPU. All training samples are generated on the fly from the GPU-resident volume, avoiding dataloader overhead. Rather than traversing the full voxel grid at each iteration \cite{liu2024baa}, training used only $N_{batch} * B$ sampled voxels per iteration, reducing per-iteration computation. 
Both interpolated and discrete training samples were generated directly on the GPU using CUDA-accelerated trilinear interpolation and direct tensor indexing. Together, on-the-fly GPU sampling and reduced host-device communication alongside CUDA-optimized kernels enabled low-latency training and inference on a single consumer-grade GPU. \pr{These steps differentiate OCTN from previously reported sparse-view INR solutions in terms of training speed by providing realistic learned representation within seconds and reducing the time barrier for integration with simulation and robotic platforms. }

\subsection{Evaluation Metrics}\label{sec:eval_metrics}

As OCT volumes contain device-dependent signal noise, attenuation, and artifacts, a diverse set of metrics is needed to adequately capture the reconstruction quality. We propose a set of local metrics, alongside global counterparts, suited for imaging modalities such as OCT, where the noise is inherent and accompanied by the global features. The local metrics quantify OCTN performance on regions of interest, while the global metrics focus on average performance over the entire image (including speckle noise), aligning with commonly reported metrics in literature for comparison.%

\subsubsection{2D image-based metrics} 
\noindent\textbf{Signal-to-noise ratio (SNR):} 
In an OCT image, the tissue regions are treated as signal, and air regions above the tissue surface are treated as noise \cite{wu2021noise}. SNR is computed \textit{per} image; therefore, a signal and noise pair in the ground-truth image is used, and similarly for the neural prediction and interpolation-generated images. In this study, representative regions \cite{ma2018speckle} with sharp intensity transitions are used for evaluation.

\noindent\textbf{Peak signal to noise ratio (PSNR)}: PSNR is computed over the same manually selected regions of interest used for SNR.

\noindent\textbf{Edge preservation index (EPI):}
EPI evaluates how well local edge structures are preserved relative to the ground-truth image \cite{ma2018speckle}. Unlike the preceding metrics, EPI is a reference-based metric. \pr{Image-based metrics are evaluated at two scales: globally over the entire generated images and locally within selected regions. In contrast, 3D metrics are reported only at the global scale.}

\subsubsection{3D surface similarity metrics}
Spatial consistency is evaluated by comparing tissue surfaces extracted from reconstructed and ground-truth volumes as pointcloud. The tissue surface is defined as the air--tissue interface, which is assumed to correspond to the largest axial intensity transition within each A-scan. For an OCT intensity field $f(x,y,z)$, with $z$ denoting the axial direction, the surface location is defined as
\begin{equation} 
\label{eq:surface_gradient_argmax}
z^{*}(x,y) = \arg\max_{z} \left| \nabla_{z} f(x,y,z) \right|,
\end{equation}
where $\nabla_z$ denotes the axial intensity gradient. For \pr{baseline native} discrete OCT volumes, $\nabla_z$ is approximated using a first-order finite difference along the axial direction.

Ground-truth and predicted surfaces are represented as point clouds, denoted by
$\mathbf{G}$ and $\mathbf{P}$, respectively. For computational efficiency, both point clouds are uniformly downsampled to at most \pr{$M=100{,}000$ points}. The 3D surface metrics are defined as follows.

\noindent\textbf{Chamfer distance (CD):}
CD quantifies the average bidirectional nearest-neighbor distance between two point clouds \cite{qi2017pointnet}. It is defined as
\begin{equation}
d_{\mathrm{CD}}(\mathbf{G},\mathbf{P}) =
\frac{1}{2N_G}\sum_{i=1}^{N_G}\min_j \|\mathbf{g}_i-\mathbf{p}_j\|_2
+
\frac{1}{2N_P}\sum_{j=1}^{N_P}\min_i \|\mathbf{p}_j-\mathbf{g}_i\|_2
\label{eq:chamfer}
\end{equation}

\noindent\textbf{Hausdorff distance (HD):}
HD measures the maximum nearest-neighbor discrepancy between two point clouds. It is defined as
\begin{equation}
d_{\mathrm{H}}(\mathbf{G},\mathbf{P}) =
\max\!\left\{
\max_i \min_j \|\mathbf{g}_i-\mathbf{p}_j\|_2,\;
\max_j \min_i \|\mathbf{p}_j-\mathbf{g}_i\|_2
\right\}.
\label{eq:hausdorff}
\end{equation}

where $\|\cdot\|_2$ denotes the Euclidean norm. Lower CD and HD values indicate greater geometric agreement, with HD capturing the worst-case surface deviation. As with the image-based metrics, both surface metrics can be affected by OCT speckle noise and a lack of distinguishing features.

\subsection{Intermediate and Sparse Reconstruction Protocol} \label{sec:intermediate_and_sparse_slice_eval_protocol}
To evaluate whether OCTN learns an isotropic continuous representation, the trained field was queried at fractional locations between adjacent acquired B-scans. 
\pr{For a volume with $D$ B-scans, anchor slices were selected around the volume center: $\left(\frac{D}{2}, \frac{D}{2}+1\right)$ for even $D$, and $\left(\frac{D-1}{2}, \frac{D+1}{2}\right)$ for odd $D$.}
Intermediate slices were rendered at $0.25$ cross-sectional intervals between the anchors. Since physical OCT measurements are not available at these fractional locations, trilinear interpolation of the acquired volume was used to generate proxy reference images. This evaluation was performed on a representative critically sampled TissueOCT volume \cref{sec:Data_acquisition_datasets}.

The protocol was then extended to sparse reconstruction from lower-density B-scan stacks. This evaluates whether OCTN can recover held-out cross-sections from sparsely acquired volumes, similar to interpolation without the need for any other method. This would help reduce acquisition burden without requiring specialized high-speed OCT hardware \cite{draelos2021contactless}. 
A subsampling factor, $SF$, was used to define the B-scan sparsity. For each $SF$, the original volume was split into acquired training B-scans and held-out ground-truth test B-scans. OCTN was trained (\cref{alg:training}) using only the acquired B-scans (\cref{fig:model_architecture}b). After training, the checkpoint was frozen and queried at the held-out B-scan locations. The reconstructed B-scans were then compared directly with corresponding ground-truth OCT measurements for fair evaluation.

\subsection{Intensity-based Volumetric Query Benchmark} \label{sec:query_speed_method}
With image and volume reconstruction tasks in place, we evaluated OCTN for robotic tasks. A simple yet common problem involved volumetric queries for an intensity-based extraction task. Each query consisted of loading the volume, filtering voxels within a specified intensity range, and applying a geometric mask to retain points within a predefined region. Three intensity ranges $[0.2-0.4]$, $[0.4-0.6]$, $[0.6-1.0]$, were used to evaluate low-,intermediate-, and high-intensity structures, respectively. \pr{Intensity range $[0.6-1.0]$ preserved high contrast tissue surface features.} For lower and upper intensity bounds, $I_{low}$ and $I_{high}$, the queried set was given 
\begin{equation}
\label{eq:intensity_query}
\mathcal{Q}_{[I_{low},I_{high}]}=\{(x,y,z)\in\mathbb{R}^3 \mid I_l \leq f(x,y,z) \leq I_h\}.
\end{equation}
Baseline OCT ground-truth volumes were stored in Neuroimaging Informatics Technology Initiative (NIFTI) file format, while for OCTN the learned volume was queried.

\subsection{Surface-Constrained Robotic Raster Path Planning} \label{sec:surface_constrained_depth_projection_method}
The continuous representation learned by OCTN enables new image-guided robotic frontiers through surface-aware path planning for OCT-guided robotic surgery. %
Consider tumor ablation with a precision robotic laser scalpel, where the laser is applied over a target region using a raster trajectory. A key requirement is to maintain a fixed focal offset from the tissue surface while preserving smooth robot motion. We therefore formulate raster execution as a surface-constrained path planning problem. %
Given a tissue volume, a raster path is first defined as $G = \{(x_i, y_i)\}_{i=1}^{N}$,
where each $(x_i, y_i)$ denotes a planned lateral location. The objective is to estimate the corresponding surface depth $z_i$ such that each point $p_i = (x_i, y_i,z_i)$ lies on, or at a prescribed offset from the surface.  We describe the steps involved in the optimization process. 

\subsubsection{Surface discovery} Tissue geometry is defined as an intensity level set,
\begin{equation} \label{eq:tissue_iso_surface}
S_{\tau} = \left\{ (x,y,z) \mid f_{\theta}(x,y,z) = \tau \right\},
\end{equation}
where $\tau$ denotes the target surface intensity level. In this study, $\tau$ is estimated from a dense-volume surface extracted using the axial-gradient method. Specifically, $\tau$ is assigned as the median intensity of the extracted surface points, providing a simple data-driven threshold for querying the learned field. 

\subsubsection{Task setup} \label{sec:raster_planning_task_setup}
A benefit of OCTN representation is the continuity of voxel space, enabling gradient-based optimization of raster points onto the tissue surface. However, direct optimization is sensitive to initialization. Since the raster path $G$ is defined only in the lateral plane, each depth value $z_i$ must be initialized near the surface. Points initialized in air have near-zero intensity gradients and can therefore fail to move toward the tissue boundary. 

We address this using a two-step strategy, and \pr{demonstrate using two separate methods, Newton, and SGD, as described below with varying benefits}. First, a coarse axial search initializes each point near the surface by sampling $N_{\mathrm{warmup}}$ candidate depths and selecting the location with the maximum finite-difference gradient magnitude. Second, the initialized points are refined using gradient-based optimization.
\begin{equation} \label{eq:step1_coarse_scan}
z_i^{(0)} = \arg\max_{z \in \mathcal{Z}} \left| \frac{\partial f_{\theta}(x_i,y_i,z)}{\partial z} \right|.
\end{equation}

\subsubsection{Newton refinement} To obtain sub-voxel surface estimates, the coarse depth initialization is refined using Newton's method. Surface discovery for $G$ is formulated as a root-finding problem along the axial direction:  
\begin{equation} \label{eq:newton_root_finding}
f_{\theta}(G_i,z) - \tau = 0
\end{equation}
Starting from coarse estimates, each depth value is updated as
\begin{equation} \label{eq:newton_updates}
z_i^{(t+1)} = z_i^{(t)} - \frac{f_{\theta}(G_i,z_i^{(t)}) - \tau}
{\frac{\partial f_{\theta}}{\partial z}(G_i,z_i^{(t)})}.
\end{equation}
In implementation, the axial gradient is approximated using a central finite difference method:
\begin{equation} \label{eq:central_difference}
\frac{\partial f_{\theta}}{\partial z}(G_i,z_i)
\approx
\frac{f_{\theta}(G_i,z_i+\epsilon) - f_{\theta}(G_i,z_i-\epsilon)}{2\epsilon},
\end{equation}

where $\epsilon$ is set to half of the OCT axial resolution in canonical coordinates. Safeguards are applied to improve stability. Updates are skipped when the gradient magnitude is too small, Newton step is clamped, and refinement is restricted to a neighborhood around the coarse estimate.

\subsubsection{SGD refinement}For uneven tissue surfaces, a smooth raster trajectory is desired to reduce robot jerk while maintaining contact with the neural iso-surface. We therefore use stochastic gradient descent-based optimization to jointly refine the raster depths and enforce axial smoothness. The surface loss penalizes deviation from the target iso-intensity level: 
\begin{equation} \label{eq:loss_surface_sgd}
\mathcal{L}_{\mathrm{surf}}(\mathbf{z}) = \sum_{i=1}^{N} \left( f_{\theta}(G_i,z_i) - \tau \right)^2,
\end{equation}
where $\mathbf{z}=[z_1,\dots,z_N]^{\top}$.

A second-order smoothness term penalizes abrupt depth changes along each raster line:
\begin{equation} 
\label{eq:loss_smooth_sgd}
\mathcal{L}_{\mathrm{smooth}}(\mathbf{z}) =
\sum_{i=2}^{N-1} v_i \left( z_{i+1}-2z_i+z_{i-1} \right)^2,
\end{equation}
where $v_i$ is a binary mask that prevents smoothing across raster-line turns, preserving sharp corners.

Together, the joint objective for the fine adjustment of raster points is given by 
\begin{equation} \label{eq:joint_objective_sgd}
\mathcal{L}(\mathbf{z}) = \mathcal{L}_{\mathrm{surf}}(\mathbf{z}) + \lambda \mathcal{L}_{\mathrm{smooth}}(\mathbf{z}),
\end{equation}
where $\lambda$ is the smoothness weight. Setting $\lambda=0$ removes smoothness regularization. During optimization, each update is clamped to remain within $N_{\mathrm{voxels}}$ of the coarse initialization, preventing points from drifting away from the local surface.

\subsubsection{Baseline} Baseline method utilizes the raw OCT voxel volume $I(D,H,W)$. For each C-scan, an initial surface is estimated through identifying the highest axial intensity change as described above. With \pr{$\mathcal{S}=\{\mathbf{s}_j\}_{j=1}^{N}, 
\quad \mathbf{s}_j=(x_j,y_j,z_j)$} surface obtained, each raster query $G$ is assigned a corresponding $z_i$ from the nearest surface point in the $XY$ plane using the k-nearest neighbor (kNN) algorithm. Due to the discrete nature of volume, depth precision is limited by axial resolution, and doing a kNN search is computationally expensive as the number of raster query points increases. \pr{Timing comparisons include only the core operations required by each method; volume or model loading and auxiliary outlier checking/rejection were excluded. In practical deployment, OCTN's speed advantage compounds over repeated high-volume spatial query tasks.}

\section{Experiments and Results}

In this section, we evaluate OCTN as a continuous volumetric representation for OCT data. We demonstrate that its ability to support spatially consistent geometric reasoning and efficient volumetric querying while allowing sparse-acquisition reconstruction as a route for faster imaging, and finally quantify cross-sectional reconstruction quality alongside comparison and ablation studies.

\subsection{OCTN Enables Fast Volumetric Query}

Imaging tissues at high resolution entails the computational cost of querying tens of millions of voxels. Conventional lookup table-based processing is discrete and time-cost increases as volume size grows \cite{liang2017ispeed}. In contrast, the OCTN learned volumes can be queried efficiently on the GPU. Here we benchmark the intensity based volumetric query-speed for a porcine muscle tissue obtained from the TissueOCT dataset  \cref{eq:neural_field_basic}, as described in \cref{sec:query_speed_method}. \pr{Reported query times included initialization, voxel filtering, geometric masking, and GPU-to-CPU transfer.}

\begin{figure}[!tb]
    \centering
    \includegraphics[width=\linewidth]{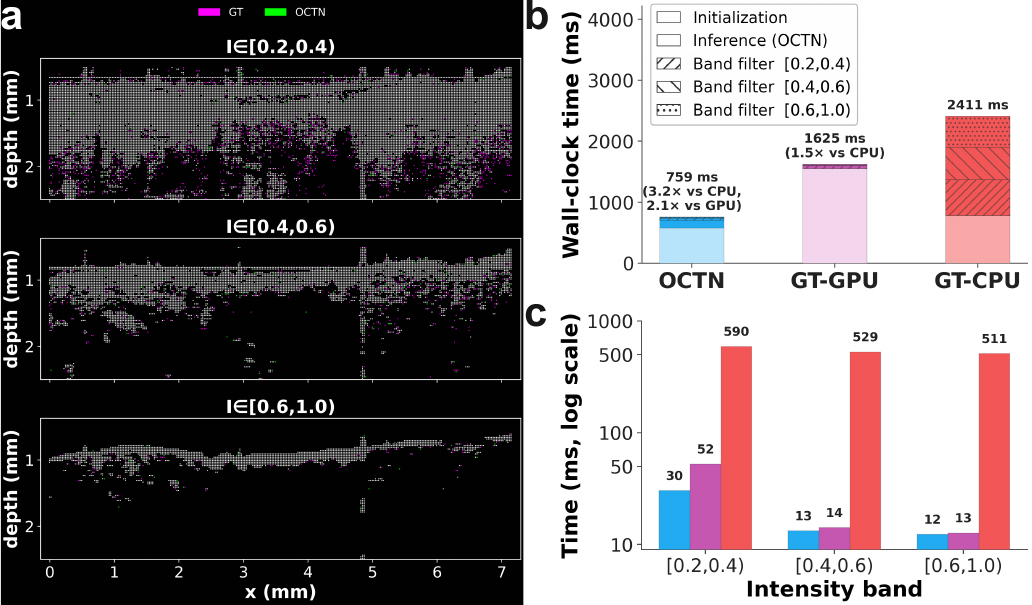}
\caption{\textbf{Intensity-based volumetric query.}
\textbf{a)} Baseline and OCTN tissue volumes across intensity ranges, with better agreement under reduced speckle.
\textbf{b)} End-to-end runtime shows $2\times$--$3\times$ OCTN speedup.
\textbf{c)} Query-only runtime shows larger gains at higher voxel counts.}
    \label{fig:fast_query}
\end{figure}

At native resolution, the corresponding OCT volume contains $256 \times 512 \times 512 = 67{,}108{,}864$ voxels. %
Our method generates qualitatively identical cross-sectional images at the tested volumetric conditions, \cref{fig:fast_query}a. This is further backed by \cref{tab:intensity_query_geometry} with point clouds generated through the learned method staying below $42\,\mu\mathrm{m}$ chamfer distance, showing excellent reproducibility of volumes and within the acceptable clinical threshold of $<0.5\,\mathrm{mm}$ for error. Hausdorff distance decreases as the points become sparse, \pr{reducing outliers in volumes}. 
It takes OCTN $0.158$~s to infer $\sim67$~M points with additional $0.0243$~s to build the uniform grid at native resolution and convert from canonical space to physical coordinate space,\cref{fig:fast_query}b, showing \pr{OCTN to have} $2\times$ speed gain over GPU and $3\times$ speed gain over CPU \pr{during the complete} process. Isolating the time needed to perform volumetric queries, OCTN achieves the target at $\sim19\times$, $\sim40\times$, $\sim43\times$ faster than CPU-based methods.  For fair comparison, the native OCT volume was transferred to GPU before evaluation. The gains over GT-GPU increase with an increase in total points queried. This is explained by an increase in GPU utilization in conventional search with an increase in net points, whereas OCTN inference batch GPU utilization remains unchanged.

\subsection{Evaluation of OCTN-guided Robot Path Planning}

\begin{figure*}[!tb]
    \centering
    \includegraphics[width=\linewidth]{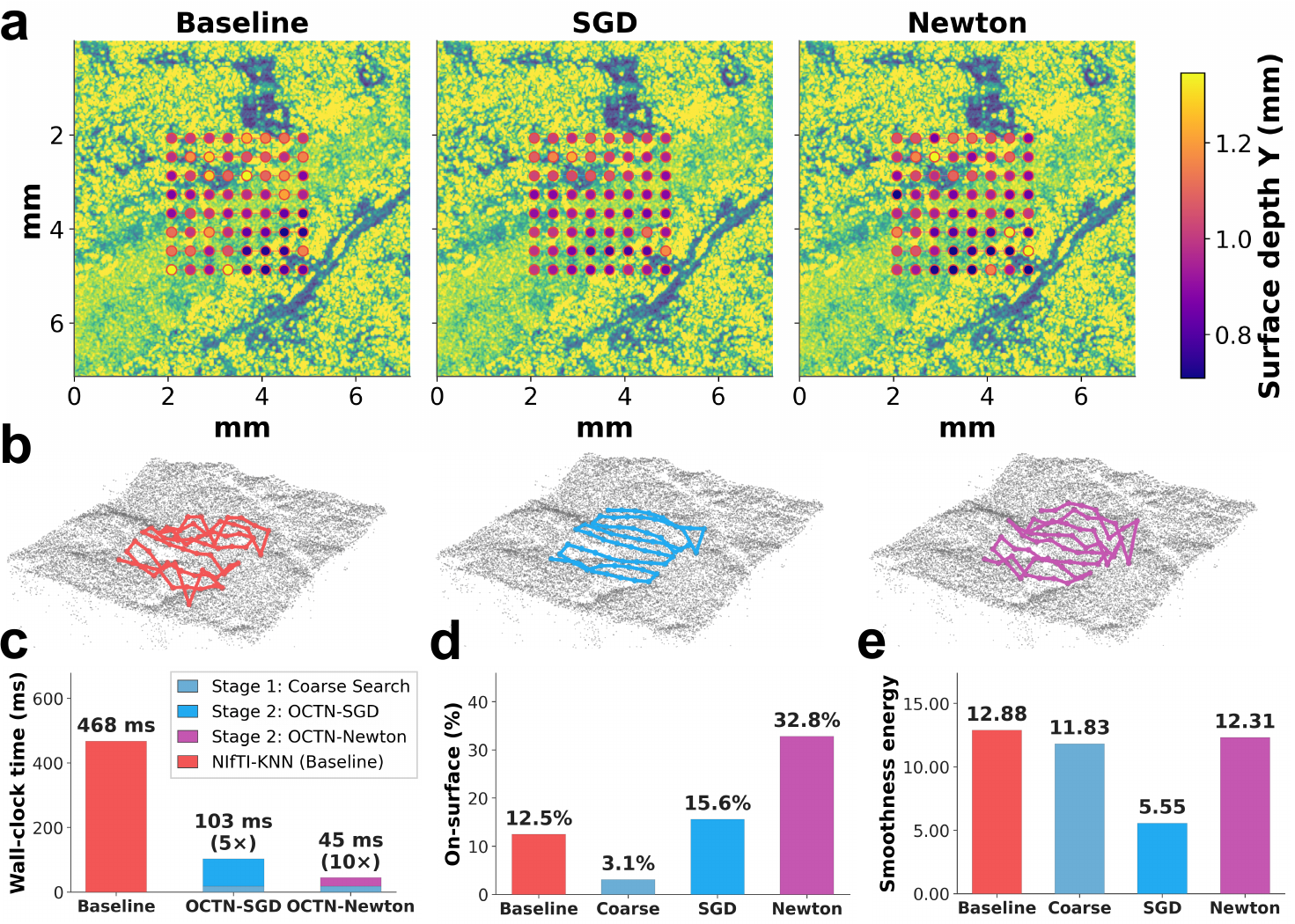}
    \caption{\textbf{OCTN-enabled path planning.} \textbf{a)} Porcine tissue MIP with a $\mathbf{3\,mm\times3\,mm}$ raster pattern; waypoints are colored by surface depth. \textbf{b)} Tissue surface with optimized raster paths. SGD enforces smoothness and reduces jerk, unlike Baseline and Newton. Methods are compared by \textbf{c)} path generation time, \textbf{d)} surface-hit rate within $5\%$ of target $\tau$, and \textbf{e)} depth smoothness energy $\Sigma(\Delta^2 y_i)^2$ ($\mathbf{mm^2}$).}
\label{fig:OCTN_raster_optimization}
\end{figure*}

We further demonstrate novel tasks towards robotic integration and guidance enabled by OCTN for the task of OCT-guided tissue resection. As described in \cref{sec:surface_constrained_depth_projection_method}, we use the RATS platform, an image-guided robotic laser surgery system with coaxial OCT imaging and laser ablation \cite{prakash2025see}, shown in \cref{fig:system_overview}b.
Here, we evaluated OCTN for two separate tasks: i) ablate a tumor $3\times3 mm^2$ in a raster pattern (\textbf{Newton method}) and ii) ensure smoothness of robot motion (\textbf{stochastic gradient descent}), as described in \cref{sec:surface_constrained_depth_projection_method}. The overall trajectory is represented in \cref{fig:OCTN_raster_optimization}a with a depth-based color map of raster points over MIP \textit{en face} image of the porcine tissue surface. %
The optimized raster path on the implicitly estimated tissue surface is shown in \cref{fig:OCTN_raster_optimization}b.
Compared to baseline kNN-based search, Newton method found the surface for the raster path $10\times$ faster, while placing $32.8\%$ points on the surface, compared to $12.5 \%$ for baseline, \cref{fig:OCTN_raster_optimization}c and  \cref{fig:OCTN_raster_optimization}d. Given relatively rough tissue surfaces and OCT speckle noise, this creates difficulties in generating smooth robot motion. The SGD-based method achieves the smoothest path with $56.9\%$ reduction over baseline, \cref{fig:OCTN_raster_optimization}e. while requiring $\sim5\times$ lower computational speed and $15.6\%$ points on the surface. 

\begin{table}[!tb]
\centering
\caption{Volumetric query accuracy across OCT intensity bands.}
 \vspace{-5pt}
\label{tab:intensity_query_geometry}
\begin{tabular}{|c|c|c|}
\hline
\textbf{Band} & \textbf{Chamfer (mm)$\downarrow$} & \textbf{Hausdorff (mm)$\downarrow$} \\
\hline
$[0.2, 0.4)$ & $0.0423$ & $1.572$ \\
$[0.4, 0.6)$ & $0.0264$ & $1.451$ \\
$[0.6, 1.0]$ & $0.0158$ & $0.963$ \\
\hline
\end{tabular}
 \vspace{-10pt}
\end{table}

\subsection{Sparse Reconstruction for Faster Volume Acquisition}
We next evaluate whether OCTN can recover dense volumetric structure from sparsely acquired B-scans, thereby reducing the number of samples required per C-scan, as per \cref{sec:intermediate_and_sparse_slice_eval_protocol}.
Volumes from TissueOCT data  were uniformly subsampled with $SF \in \{2,4,8,16,32\}$, and OCTN was trained using only the retained training volume part. Reconstructed B-scans were evaluated against held-out ground truth B-scans. Trilinear interpolation was used as baseline.

\begin{figure*}[!tb]
    \centering
    \includegraphics[width=\linewidth]{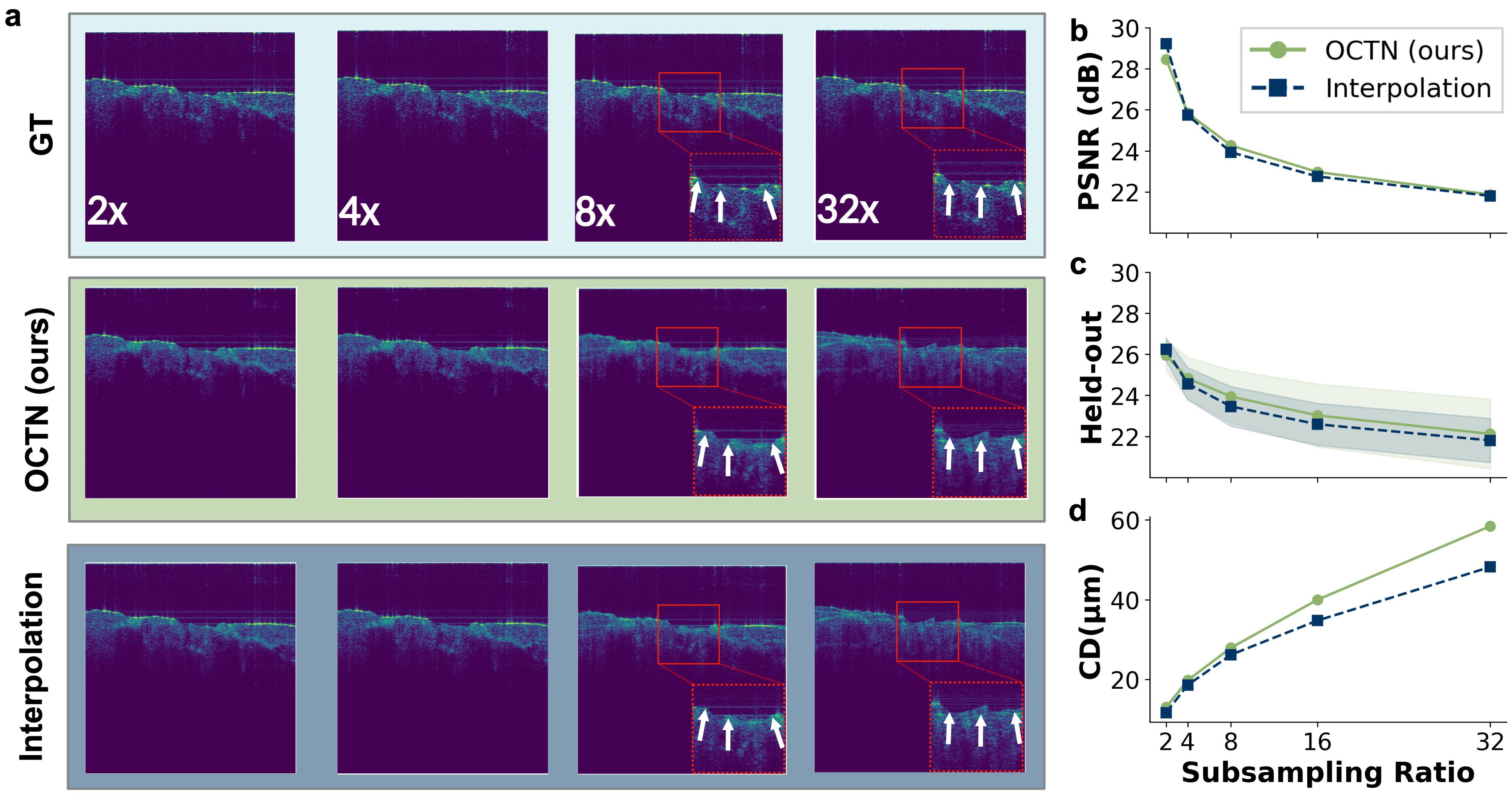}

\caption{\textbf{Sparse volume reconstruction.} \textbf{a)} Qualitative B-scan comparison across $SF$ for OCTN and interpolation. Red boxes and white arrows mark interpolation-induced surface artifacts visible in $8\times$ and beyond. \textbf{b)} Global PSNR at native resolution $\left[256,512,512\right]$. \textbf{c)} Held-out B-scan performance reported as mean $\pm 1\sigma$ over each volume. \textbf{d)} Chamfer distance (CD) quantifies reconstruction quality and spatial-feature preservation.}
    \label{fig:subsampling_images}
\end{figure*}

OCTN consistently matched or exceeded trilinear interpolation on held-out B-scans. For qualitative comparison \cref{fig:subsampling_images}a shows representative slices selected near the midpoint between neighboring acquired B-scans, where interpolation is least favored but the evaluation is most informative. At aggressive subsampling levels, beginning at $SF=8\times$, \pr{with inter B-scan distance of $224\,\mu\mathrm{m}$,} visible blurring of fine tissue features is observed (highlighted in red), with substantial degradation seen at $SF=32\times$. This is expected: at $SF=8\times$, only $32$ of $256$ B-scans are used for training, and at $SF=32\times$, only $8$ of $256$ B-scans are available \pr{at $896\,\mu\mathrm{m}$ spacing between acquired data}. Consistent with sampling limits, neither OCTN nor interpolation can recover high-frequency structure that is absent from the acquired data.
We evaluated the trained neural representation on the full grid at native resolution, $\left[256,\,512,\,512\right]$, as shown in \cref{fig:subsampling_images}b, and at held-out dataset in \cref{fig:subsampling_images}c.  As a baseline, trilinear interpolation is used to upsample the subsampled input volume from  $\left[256/SF,\,512,\,512\right]$ to $\left[256,\,512,\,512\right]$. OCTN demonstrates improved performance over standard method without the extra step needed for interpolation.
Furthermore, as seen in \cref{fig:subsampling_images}d, volumes reconstructed through OCTN shows closeness to ground truth data. OCTN prediction provides smoother representation even for conditions with higher $SF$.

\subsection{Reconstruction of OCT B-scans}

\begin{table*}[!tb]
\centering
\caption{Global and local reconstruction performance across representative datasets. $\Delta m = m_{\mathrm{ROI}} - m_{\mathrm{global}}$.}
\label{tab:roi_reconstruction_all_datasets}
\scriptsize
\setlength{\tabcolsep}{2.0pt}
\renewcommand{\arraystretch}{1.12}
\begin{tabular}{|c|c|c|c|c|c|c|c|c|}
\hline
\textbf{Dataset} &
\textbf{Index} &
\textbf{Global} &
\textbf{ROI} &
\textbf{ROI} &
$\boldsymbol{\Delta}$\textbf{ROI} &
$\mathbf{SNR_{GT}}$ $\uparrow$ &
$\mathbf{SNR_{Pred}}$ $\uparrow$ &
\textbf{EPI $\sim1$} \\
&
&
\textbf{PSNR $\uparrow$/ SSIM $\uparrow$} &
&
\textbf{PSNR $\uparrow$ / SSIM $\uparrow$} &
\textbf{PSNR $\uparrow$ / SSIM $\uparrow$} &
\textbf{(dB)} &
\textbf{(dB)} &
\\
\hline

\multirow{3}{*}{\textbf{Srinivasan}} 
& \multirow{3}{*}{24} 
& \multirow{3}{*}{44.64 / 0.996}
& R1 & 44.81 / 0.997 & +0.17 / +0.001 & 9.42 & 9.41 & 0.998 \\
&
&
& R2 & 43.39 / 0.992 & -1.25 / -0.004 & 3.32 & 3.32 & 0.977 \\
&
&
& R3 & 43.60 / 0.999 & -1.04 / +0.003 & 6.62 & 6.62 & 0.992 \\
\hline

\multirow{3}{*}{\textbf{Harvard}} 
& \multirow{3}{*}{100} 
& \multirow{3}{*}{45.05 / 0.995}
& R1 & 45.24 / 0.997 & +0.19 / +0.002 & 4.87 & 4.86 & 0.990 \\
&
&
& R2 & 46.09 / 0.999 & +1.04 / +0.004 & 8.47 & 8.48 & 1.000 \\
&
&
& R3 & 45.77 / 0.999 & +0.72 / +0.004 & 9.67 & 9.67 & 0.989 \\
\hline

\multirow{3}{*}{\textbf{Noor Eye}} 
& \multirow{3}{*}{12} 
& \multirow{3}{*}{49.00 / 0.997}
& R1 & 47.96 / 0.998 & -1.04 / +0.001 & 11.96 & 11.98 & 1.048 \\
&
&
& R2 & 49.00 / 0.997 & -0.00 / +0.000 & 9.84 & 9.86 & 1.058 \\
&
&
& R3 & 50.05 / 1.000 & +1.05 / +0.003 & 10.44 & 10.46 & 1.016 \\
\hline

\multirow{3}{*}{\textbf{TissueOCT}} 
& \multirow{3}{*}{128} 
& \multirow{3}{*}{35.61 / 0.833}
& R1 & 40.34 / 0.997 & +4.73 / +0.164 & 16.56 & 18.66 & 0.994 \\
&
&
& R2 & 36.99 / 0.988 & +1.38 / +0.155 & 13.56 & 15.64 & 0.981 \\
&
&
& R3 & 29.40 / 0.988 & -6.21 / +0.155 & 17.65 & 19.72 & 0.897 \\
\hline

\multirow{3}{*}{\textbf{TissueOCT Gaussian}} 
& \multirow{3}{*}{128} 
& \multirow{3}{*}{38.34 / 0.870}
& R1 & 37.43 / 0.985 & -0.91 / +0.115 & 19.72 & 22.56 & 0.955 \\
&
&
& R2 & 36.65 / 0.968 & -1.69 / +0.098 & 16.39 & 19.20 & 0.958 \\
&
&
& R3 & 33.60 / 0.992 & -4.74 / +0.122 & 20.66 & 23.52 & 0.979 \\
\hline

\multirow{3}{*}{\textbf{TissueOCT Median}} 
& \multirow{3}{*}{128} 
& \multirow{3}{*}{39.04 / 0.955}
& R1 & 36.80 / 0.984 & -2.24 / +0.029 & 22.75 & 25.63 & 1.001 \\
&
&
& R2 & 34.74 / 0.961 & -4.30 / +0.006 & 19.42 & 22.25 & 0.992 \\
&
&
& R3 & 34.61 / 0.995 & -4.43 / +0.040 & 23.57 & 26.48 & 0.965 \\
\hline

\end{tabular}
\end{table*}

\begin{figure}[!tb]
  \centering
    \includegraphics[width=0.78\linewidth,keepaspectratio]{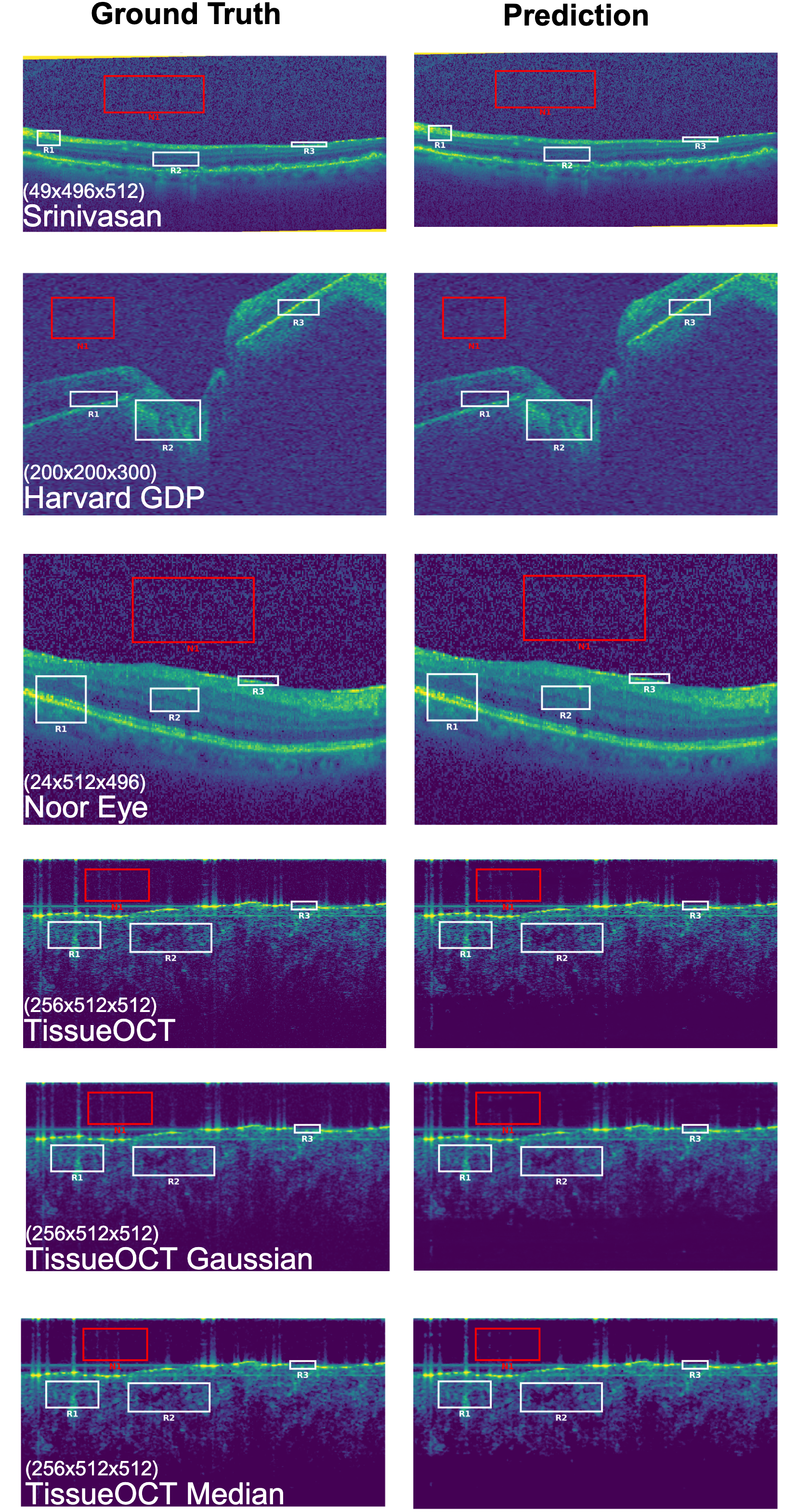}

  \caption{\textbf{B-scan reconstruction}: OCTN performance over Srinivasan \cite{srinivasan2014fully}, Harvard GDP \cite{luo2023harvard}, Noor Eye \cite{rasti2017macular}, and variants of proposed TissueOCT data. For local metric estimation, regions marked as noise are shown in red, and signal in white, with results in \cref{tab:roi_reconstruction_all_datasets}.  
  }
  \label{fig:2D_dataset_comparison}
\end{figure}

Finally, the benefits of robot integration and sparse reconstruction cannot be achieved at the cost of reduction in B-scan reconstruction quality. We assess cross-dataset generalization by evaluating OCTN over public swept-source and spectral-domain datasets (Duke Srinivasan \cite{srinivasan2014fully},  Harvard GDP \cite{luo2023harvard}, Noor Eye \cite{rasti2017macular} and our TissueOCT dataset). \rp{Two variants of TissueOCT dataset volumes are generated with smoothing (Gaussian and Median with filter of 3) to gently smooth the noise.}
\rp{For qualitative comparison, B-scan midway through the volume was chosen for evaluation. Here, $\alpha = 0.20$ was chosen throughout. Local metrics are calculated on the regions of interest, here defined by intensity change and air-tissue boundary. The results \cref{fig:2D_dataset_comparison} show OCTN is able to reconstruct volumes at high rendering quality. As seen in \cref{tab:roi_reconstruction_all_datasets}, rendered images maintain the native SNR, highlighting the ability to learn high-frequency image features. For datasets containing high speckle noise (TissueOCT), local PSNR and SSIM show gains for $R1$ and $R2$ but experience degradation in performance at the air--tissue boundary ($R3$). \pr{Regions were selected based on local intensity-feature variation to highlight areas where reconstruction is more challenging.} TissueOCT Gaussian and Median show higher global metrics with degraded performance locally, suggesting inflation due to the smoothing of speckle noise. Presence of EPI of $\sim1$ shows the preserves edge well relative to the ground truth.} \pr{We report PSNR, SSIM, and EPI relative to regions in ground truth.}

\begin{figure*}[!tb]
  \centering
  \includegraphics[width=0.92\linewidth]{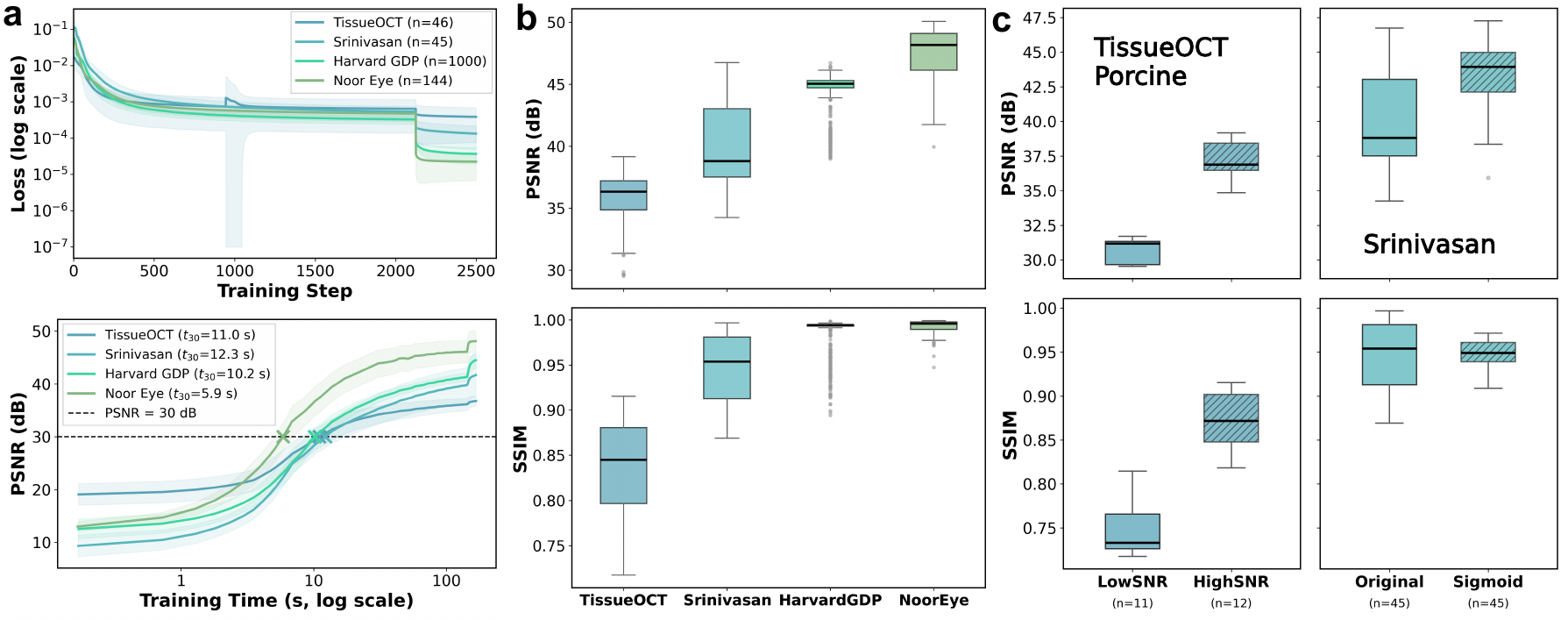}
  \caption{\textbf{OCTN performance across datasets.} \textbf{a)} Training loss and PSNR, showing two-stage training reaches $>30$~dB PSNR within seconds. PSNR is computed on 16 uniformly spaced B-scans from 10 random volumes. \textbf{b)} Global PSNR and SSIM distributions. \textbf{c)} Speckle-noise reduction improves PSNR and SSIM as seen in TissueOCT (left) and Srinivasan (right) volumes.}
  \label{fig:2d_metrics}
\end{figure*}

\rp{Furthermore, the two-stage training helps in lowering the loss and improving PSNR values, \cref{fig:2d_metrics}a, achieving acceptable PSNR ($30$~dB) in less than $10$~s on average. Reconstruction quality generally tends to improve with a reduction in total voxels in a dataset, \cref{fig:2d_metrics}b. Higher variance in performance stems from varied C-scan dimensions and noise levels within datasets. To highlight the effect of speckle noise on rendering performance beyond TissueOCT dataset, a noise-suppressing step is applied on the Srinivasan dataset. Results, as seen in \cref{fig:2d_metrics}c, highlight improvement of $6.5$~dB and $3.3$~dB in PSNR while $0.12$ and $0.002$ in SSIM, respectively, for TissueOCT (Porcine) and Srinivasan datasets, showing effectiveness.}

\begin{figure*}[!tb]
    \centering
    \includegraphics[width=0.92\linewidth]{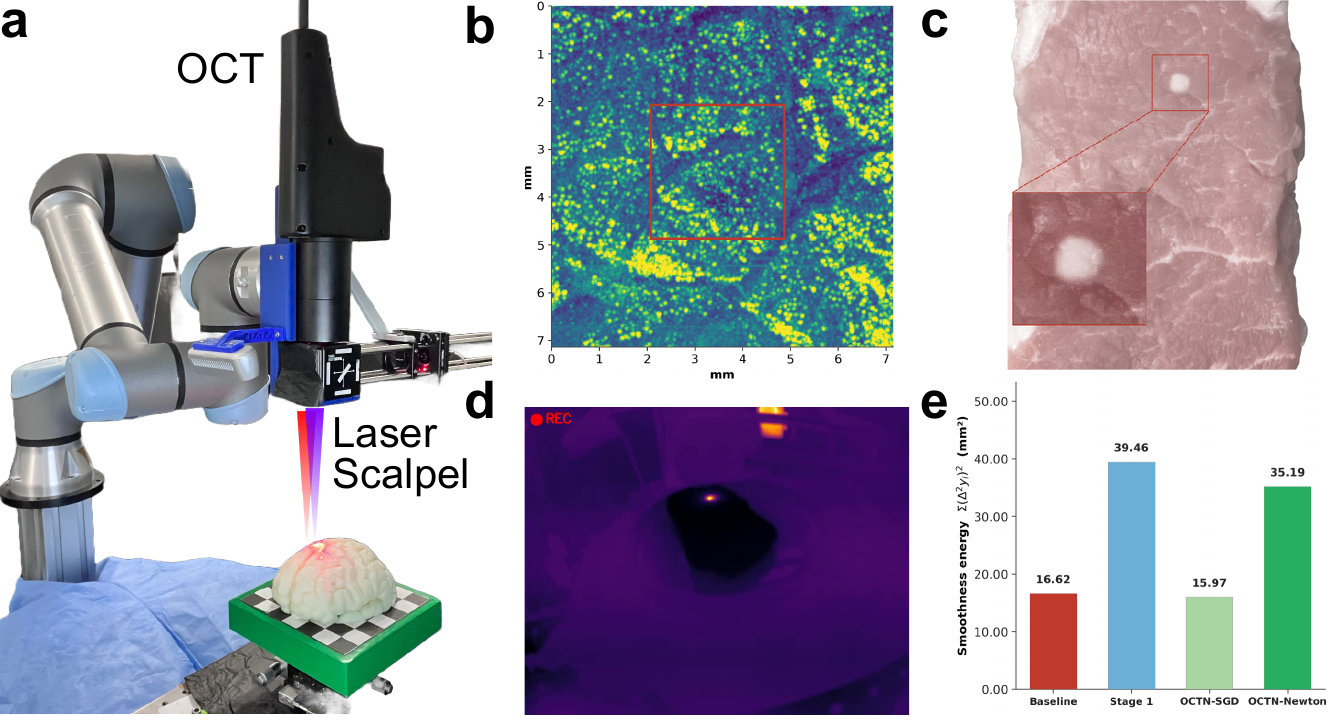}
\caption{\pr{\textbf{Porcine tissue ablation demonstration.} \textbf{a)} RATS robotic laser platform integrating OCT and a laser scalpel. \textbf{b)} \textit{En face} MIP of porcine with target region marked in red. \textbf{c)} Post-ablation tissue showing localized coagulation and discoloration, with corresponding \textbf{d)} thermal image. \textbf{e)} Smoothness comparison, showing SGD achieves the lowest smoothness energy in the physical experiment.}}
    \label{fig:ablation_demo}
\end{figure*}

\subsection{Real World \textit{Ex vivo} Ablation}
\rp{We perform real-world tissue ablation experiments to translate the properties of OCTN to a physical robotic surgery platform as described in \cref{sec:surface_constrained_depth_projection_method}. RATS laser surgery platform is used, \cref{fig:ablation_demo}a, \cite{prakash2025see} with the task of tissue ablation in a $3 mm \times 3mm$ region. Porcine tissue surface was captured at native resolution, \pr{as seen in \cref{fig:ablation_demo}b}. OCTN was used to learn a neural representation of volume within \pr{$1000$ iterations at $\alpha$ $0.20$}, and raster waypoints were generated, using Newton, SGD ($\lambda= 10$ \pr{tuned based on tissue noise}), and baseline methods. }
\rp{Using methods described in \cite{prakash2025see}, a coordinate transformation matrix from OCT physical frame to Robot end effector (EE) frame was obtained,$T^{EE}_{OCT}$, allowing conversion of $T^{World}_{Tissue}$ to the frame. }\pr{Qualitatively, ablation was observed, as seen in \cref{fig:ablation_demo}c with discoloration and concurrently in, as seen in \cref{fig:ablation_demo}d in the thermal image.} During implementation, OCTN helped by reducing the computational cost of generating waypoints and reducing-jerk robot motion \pr{\cref{fig:ablation_demo}e.}

\subsection{Ablation Studies}
We test the robustness of the proposed methods through ablation studies. \rp{First, the effect of $
\alpha$ in the two-stage training strategy is evaluated through global PSNR/SSIM metric on volume at native discrete pixel locations, and at a fractional B-scan at position $128.5$. The value is chosen around the center of the volume. As shown in \cref{tab:octn_ablation}, with increasing $alpha$, the global discrete metrics decrease while the fractional metrics increase. The ideal value of $\alpha= 0.20$ provides highest PSNR at discrete locations with marginal degradation of image quality at fractional locations. The gain of $4.33$~dB in PSNR and $0.161$ in SSIM on average from $\alpha= 0.0$ to $\alpha= 0.20$ shows the effect that marginal involvement of interpolated voxels during training can have.} %
\rp{For comparison, we chose CuNeRF~\cite{chen2023cunerf}, an INR method for CT and MRI which follows a similar input and output space. The hyperparameters were author-reported, with minor modifications made to training and evaluation batch size pertaining to the GPU in use. As seen in \cref{tab:octn_cunerf_comparison}, OCTN consistently performs better than CuNeRF with the final reconstruction losing sharpness for high intensity features. The extended training time $> 6$~hrs makes it difficult for use with OCT, where each volume represents a minuscule area of $7 \times 7$ $mm^2$.  \pr{Reported results were evaluated on uniformly sampled B-scans for both the methods.}CuNeRF may find benefit with modalities such as CT and MRI with volumes representing larger tissue regions, though OCTN might show an advantage in terms of speed.}

\begin{table}[!tb]
\centering
\caption{Ablation of OCTN two-stage training on TissueOCT. Metrics are reported for the full volume and fractional B-scan 128.5.}
\vspace{-5pt}
\label{tab:octn_ablation}
\scriptsize
\setlength{\tabcolsep}{2.2pt}
\renewcommand{\arraystretch}{1.08}
\begin{tabularx}{\columnwidth}{|>{\raggedright\arraybackslash}X|c|c|c|}
\hline
\textbf{Strategy} &
$\boldsymbol{\alpha}$ &
\textbf{Volume PSNR/SSIM $\uparrow$} &
\textbf{128.5 PSNR/SSIM $\uparrow$} \\
\hline
Discrete only
& $0.00$
& $35.26{\pm}1.44$ / $\mathbf{0.87{\pm}0.03}$
& $22.86{\pm}3.19$ / $0.506{\pm}0.192$ \\
\hline
OCTN, proposed
& $\mathbf{0.20}$
& $\mathbf{35.35{\pm}2.87}$ / $0.83{\pm}0.06$
& $27.19{\pm}3.12$ / $0.667{\pm}0.160$ \\
\hline
Low-interp.
& $0.40$
& $34.93{\pm}3.30$ / $0.81{\pm}0.08$
& $28.04{\pm}3.00$ / $0.698{\pm}0.144$ \\
\hline
Moderate-interp.
& $0.60$
& $34.32{\pm}3.50$ / $0.79{\pm}0.09$
& $28.55{\pm}2.89$ / $0.713{\pm}0.134$ \\
\hline
High-interp.
& $0.80$
& $33.62{\pm}3.48$ / $0.80{\pm}0.03$
& $\mathbf{28.79{\pm}2.85}$ / $\mathbf{0.723{\pm}0.128}$ \\
\hline
Interp. only
& $1.00$
& $32.74{\pm}3.34$ / $0.75{\pm}0.10$
& $28.20{\pm}2.63$ / $0.716{\pm}0.120$ \\
\hline
\end{tabularx}
\vspace{-10pt}
\end{table}

\begin{table}[!tb]
\centering
\caption{Comparison of OCTN and CuNeRF \cite{chen2023cunerf} ($n=3$): $\sim171$ s/2.5k steps versus $\sim394$ min/250k steps.}
\label{tab:octn_cunerf_comparison}
\scriptsize
\setlength{\tabcolsep}{3.0pt}
\renewcommand{\arraystretch}{1.08}
\begin{tabular}{|p{0.34\linewidth}|c|c|c|}
\hline
\textbf{Dataset} & \textbf{Method} & \textbf{PSNR $\uparrow$} & \textbf{SSIM $\uparrow$} \\
\hline

\multirow{2}{=}{Srinivasan \cite{srinivasan2014fully}}
& CuNeRF & $12.98 \pm 0.96$ & $0.38 \pm 0.09$ \\
& OCTN & $\mathbf{41.99 \pm 5.33}$ & $\mathbf{0.97 \pm 0.025}$ \\
\hline

\multirow{2}{=}{Harvard GDP \cite{luo2023harvard}}
& CuNeRF & $24.70 \pm 0.78$ & $0.48 \pm 0.06$ \\
& OCTN & $\mathbf{44.55 \pm 1.23}$ & $\mathbf{0.99 \pm 0.01}$ \\
\hline

\multirow{2}{=}{Noor Eye \cite{rasti2017macular}}
& CuNeRF & $19.62 \pm 1.95$ & $0.30 \pm 0.038$ \\
& OCTN & $\mathbf{45.95 \pm 1.27}$ & $\mathbf{0.99 \pm 0.00}$ \\
\hline

\multirow{2}{=}{TissueOCT (ours)}
& CuNeRF & $23.82 \pm 0.93$ & $0.69 \pm 0.01$ \\
& OCTN & $\mathbf{35.09 \pm 0.32}$ & $\mathbf{0.84 \pm 0.02}$ \\
\hline
\end{tabular}
\vspace{-15pt}
\end{table}

\section{Discussion}\label{sec:discussion}
Widespread adoption of OCT for intraoperative surgical guidance is hindered by imaging anisotropy, acquisition speed, and computational complexity involved in data handling. Here, we present OCTN, an isotropic continuous representation of OCT volumes allowing up to $4\times$ reduction in acquisition speed and fast volumetric query ($\sim67$~M voxels in $0.158$~s). With optimized two-stage training strategy, OCTN allows learning tissue volumes within seconds ($>30$~PSNR in $<10$~s) and permit sparse reconstruction of tissue volumes. In contrast with recent INR works in OCT utilizing external sensors \cite{kahrs2026don} or theoretical model \cite{li2025computational} of OCT, OCTN is hardware agnostic, and develops a direct map between spatial coordinates and voxel intensity by learning a continuous smooth volume. When coupled with a dexterous, precise laser scalpel, OCTN enables smooth tissue ablation by optimizing with sub-voxel accuracy at near-real-time speed. a %

Overall, we evaluated OCTN on two categories of tasks: i) pertaining to current clinical diagnostics and ii) enabling integration with robotic hardware for guiding tools. 
For the first part, the core contribution is the hybrid training strategy. This allows reconstruction of B-scans in plane of choice, as desired by the clinicians. The use of local image-based and 3D surface similarity metrics perform holistic evaluation while the global metrics allow easy comparison with popular methods on multiple datasets. Downstream robotic integration and fast queries are the byproducts of the learned continuous volume residing in the GPU. This allowed us to perform intensity-based volumetric queries at $3.3\times$ faster than on CPU and about $2.1\times$ faster than on GPU (includes initialization and filtering). This speed gain is visible for queries with higher computational load. For the task of tissue ablation using the RATS~\cite{prakash2025see} surgical system, we successfully generate a raster path with over $39\%$ of waypoints on the tissue surface (\textbf{Newton method}) and lower smoothness energy (\textbf{SGD method}) over baseline by formulating the surface discovery as an optimization problem with OCTN allowing easy gradient calculation. This is finally demonstrated on \textit{ex vivo} porcine tissue resulting in translation of OCTN to real world tasks. 

Despite benefits, OCTN has a few limitations. For high-resolution tissue volumes ($>250$~M voxels) obtained from the latest SS-OCTs, we observe degradation in reconstruction quality, amounting to a pseudo-Gaussian blur effect for an MLP of fixed size. Background noise is detrimental to the final learned quality, and there is no explicit method of handling it. Possible preprocessing to suppress background noise, or a specialized loss function during training, can assist in overcoming the challenge. Another issue stems from the choice of learning architecture used. Fully-fused MLP limits the backpropagation of first-order gradients during optimization with recent fixes suggesting potential solutions. Once solved, we anticipate OCTN to be the representation of choice for OCT data handling and integration.

Future work will extend OCTN toward dynamic surgical guidance by integrating clinical segmentation and diagnostic models, such as SAM2, with temporal inputs for deforming tissue. In ocular imaging, coupling OCTN with scanning laser ophthalmoscopy may improve reconstruction \cite{kahrs2026don}, while sequential OCT registration could support temporally consistent tissue representations \cite{fang2025robotic}. Deriving signed distance functions from OCTN volumes may further enable faster geometric queries and planning \cite{ma20233d,wang2025sampling}, including guidance for needles and scalpels \cite{ma2025tumormap}. Together, these directions position OCTN as a robot-ready OCT representation for multimodal surgical guidance, moving OCT beyond ophthalmic workflows toward deformable soft-tissue surgery, tumor resection, and broader sensor-fusion-driven \cite{chen2022fully} robotic intervention.

\section{Conclusion}
We present OCTN, an INR-based method for generating isotropic, continuous OCT volumes for enabling tasks in geometrical reasoning, robotics and imaging.
OCTN proposes a two-stage training strategy to utilize native discrete and interpolated voxels to convert discrete OCT C-scans into continuous neural counterparts agnostic of the OCT sensor.
With GPU-native data handling, OCTN allows enabling fast volumetric query, sparse volume reconstruction and easy out-of-the-box integration with surgical robotic platforms for path planning and tissue ablation with minimal loss in reconstruction quality. Future work would extend OCTN to include spatial and temporal sensor information on for wider clinical utility. The adjoining TissueOCT dataset and codebase are released to allow one-step conversion of OCT data to supercharged OCTN representation for public use.

\section*{REFERENCES}
 \vspace{-15pt}
\bibliographystyle{IEEEtran} 
\bibliography{bibtex}

\end{document}